\documentclass[runningheads]{llncs}

\usepackage{times}
\usepackage{subcaption}
\usepackage{graphicx}
\usepackage{multirow}
\usepackage{todonotes}
\usepackage{soulutf8}
\usepackage{soul}
\begin{document}

\title{Unmasking Bias in News}

\author{Javier S\'anchez-Junquera\inst{1}\and
Paolo Rosso\inst{1}
\and\\ Manuel Montes-y-G\'omez \inst{2} \and Simone Paolo Ponzetto\inst{3}}

\authorrunning{S\'anchez-Junquera et al.}

\institute{PRHLT Research Center, Universitat Polit\`ecnica de Val\`encia, Spain \email{jjsjunquera@gmail.com}, \email{prosso@dsic.upv.es} \and
Instituto Nacional de Astrof\'isica \'Optica y Electr\'onica, Mexico \\
\email{mmontesg@inaoep.mx}  \and
Data and Web Science Group,University of Mannheim, Germany \\
\email{simone@informatik.uni-mannheim.de }
}
\maketitle 

\begin{abstract}
  We present experiments on detecting hyperpartisanship in news using a `masking' method that allows us to assess the role of style vs.\ content for the task at hand. Our results corroborate previous research on this task in that topic related features yield better results than stylistic ones. We additionally show that competitive results can be achieved by simply including higher-length n-grams, which suggests the need to develop more challenging datasets and tasks that address implicit and more subtle forms of bias.
  
\end{abstract}

\section{Introduction}

Media such as radio, TV channels, and newspapers control which information spreads and how it does it. The aim is often not only to inform readers but also to influence public opinion on specific topics from a hyperpartisan perspective. 

Social media, in particular, have become the default channel for many people to access information and express ideas and opinions. The most relevant and positive effect is the democratization of information and knowledge but there are also undesired effects. 
One of them is that social media foster information bubbles: every user may end up receiving only the information that matches her personal biases, beliefs, tastes and points of view. Because of this, social media are a breeding ground for the  propagation of fake news: when a piece of news outrages us or matches our beliefs, we tend to share it without checking its veracity; and, on the other hand, content selection algorithms in social media give credit to this type of popularity because of the click-based economy on which their business are based. Another harmful effect is that the  relative anonymity of social networks facilitates the propagation of toxic, hate and exclusion messages. Therefore, social media  contribute to the misinformation and polarization of society, as we have recently witnessed in the last presidential elections in USA or the Brexit referendum. Clearly, the polarization of society and its underlying discourses are not limited to social media, but rather reflected also in political dynamics (e.g., like those found in the US Congress \cite{andr15-rise}): even in this domain, however, social media can provide a useful signal to estimate partisanship \cite{hemphill16}.

Closely related to the concept of controversy and the ``filter bubble   effect'' is the concept of bias \cite{Baeza-Yates2018}, which refers to the  presentation of information according to the standpoints or interests of the journalists and the news agencies. Detecting bias is very important to help users to acquire balanced information. Moreover, how a piece of information is reported has the capacity to evoke different sentiments in the audience, which may have large social implications (especially in very controversial topics such as terror attacks and religion issues).

In this paper, we approach this very broad topic by focusing on the problem of  detecting hyperpartisan news, namely news written with an extreme manipulation of the reality on the basis of an underlying, typically extreme, ideology. This problem has received little attention in the context of the automatic detection of fake news, despite the potential correlation between them. Seminal work from \cite{P18-1022} presents a comparative style analysis of hyperpartisan news, evaluating features such as characters n-grams, stop words, part-of-speech, readability scores, and ratios of quoted words and external links. The results indicate that a topic-based model outperforms a style-based one to separate the left, right and mainstream orientations.

We build upon previous work and use the dataset from \cite{P18-1022}: this way we can investigate hyperpartisan-biased news (i.e., extremely one-sided) that have been manually fact-checked by professional journalists from BuzzFeed. The articles originated from 9 well-known political publishers, three each from the mainstream, the  hyperpartisan left-wing, and the hyperpartisan right-wing. To detect hyperpartisanship, we apply a masking technique that transforms the original texts in a form where the textual structure is maintained, while letting the learning algorithm focus on the writing style or the topic-related information. This technique makes it possible for us to corroborate previous results that content matters more than style. However, perhaps surprisingly, we are able to achieve the overall best performance by simply using higher-length n-grams than those used in the original work from \cite{P18-1022}: this seems to indicate a strong lexical overlap between different sources with the same orientation, which, in turn, calls for more challenging datasets and task formulations to encourage the development of models covering more subtle, i.e., implicit, forms of bias.

\vspace{1em}
\noindent
The rest of the paper is structured as follows. In Section \ref{Masking_technique} we describe our method to hyperpartisan news detection based on masking. Section \ref{mask_hyp} presents details on the dataset, experimental results and a discussion of our results. Finally, Section \ref{conclusion} concludes with some directions  for future work.

\section{Investigating masking for hyperpartisanship detection}
\label{Masking_technique}

The masking technique that we propose here for the hyperpartisan news detection task has been applied to text clustering \cite{granados2011reducing}, authorship attribution \cite{stamatatos2017authorship}, and recently to deception detection \cite{er} with encouraging results. The main idea of the proposed method is to transform the original texts to a form where the textual
structure, related to a general style (or topic), is maintained while content-related (or style-related) words are masked. To this end, all the occurrences (in both training and test corpora) of non-desired terms are replaced by symbols.

Let $W_k$ be the set of the $k$ most frequent words, we mask all the occurrences of a word $w \in W_k$ if we want to learn a \textit{topic-related model}, or we mask all $w \notin W_k$ if we want to learn a \textit{style-based model}. Whatever the case, the way in which we mask the terms in this work is called \textit{Distorted View with Single Asterisks} and consists in replacing $w$ with a single asterisk or a single \# symbol if the term is a word or a number, respectively. For further masking methods, refer to \cite{stamatatos2017authorship}.

Table \ref{tab:example} shows a fragment of an original text and the result of masking style-related information or topic-related information. With the former we obtain distorted texts that allow for learning a \textit{topic-based model}; on the other hand, with the latter, it is possible to learn a \textit{style-based model}. One of the options to choose the terms to be masked or maintained without masking is to take the most frequent words of the target language \cite{stamatatos2017authorship}. In the original text from the table, we highlight some of the more frequent words in English. 
\begin{table}[t]
\centering
\caption{Examples of masking style-related information or topic-related information.}
\label{tab:example}
\begin{tabular}{|p{0.37\textwidth}|p{0.30\textwidth}|p{0.30\textwidth}|}
\hline
\multicolumn{1}{|c|}{\textbf{Original text}} & \multicolumn{1}{c|}{\textbf{\begin{tabular}[c]{@{}c@{}}Masking topic-\\ related words\end{tabular}}} & \multicolumn{1}{c|}{\textbf{\begin{tabular}[c]{@{}c@{}}Masking style-\\ related words\end{tabular}}} \\ \hline
Officers \textbf{went after} Christopher \textbf{Few} \textbf{after} watching \textbf{an} argument \textbf{between him and his} girlfriend outside \textbf{a} bar \textbf{just before the} 2015 shooting & * went after * Few after * an * between him and his * a  * just before the \# * & Officers * * Christopher * * watching * argument * * * * girlfriend outside * bar * * * 2015 shooting \\ \hline
\end{tabular}
\end{table}

\section{Experiments}
\label{mask_hyp}

We used the BuzzedFeed-Webis Fake News Corpus 2016 collected by \cite{P18-1022} whose articles were labeled with respect to three political orientations: mainstream, left-wing, and right-wing (see Table \ref{tab:data}). Each article was taken from one of 9 publishers known as hyperpartisan left/right or mainstream in a period close to the US presidential elections of 2016. Therefore, the content of all the articles is related to the same topic.  

\begin{table}[t]
\caption{Statistics of the original dataset and its subset used in this paper.}
\label{tab:data}
\centering
\begin{tabular}{|l|c|c|c||c|}
\hline
\multicolumn{1}{|c|}{} & \textbf{Left-wing} & \textbf{Mainstream} & \textbf{Right-wing} & \multicolumn{1}{c|}{ \textbf{\tiny $\sum$}} \\ \hline
\textbf{Original data \cite{P18-1022}} & 256 & 826 & 545 & 1627 \\ \hline
\textbf{Cleaned data} & 252 & 787 & 516 & 1555 \\ \hline 
\end{tabular}
\end{table}

During initial data analysis and prototyping we identified a variety of issues with the original dataset: we cleaned the data excluding articles with empty or bogus texts, e.g. \textit{`The document has moved here}' (23 and 14 articles respectively). Additionally, we removed duplicates (33) and files with the same text but inconsistent labels (2). As a result, we obtained a new dataset with 1555 articles out of 1627.\footnote{The dataset is available at \url{https://github.com/jjsjunquera/UnmaskingBiasInNews}.} Following the settings of \cite{P18-1022}, we balance the training set using random duplicate oversampling. 

\subsection{Masking Content vs.\ Style in Hyperpartisan News}

In this section, we reported the results of the masking technique from two different perspectives. In one setting, we masked \emph{topic-related information} in order to maintain the predominant writing style used in each orientation. We call this approach a \emph{style-based model}. With that intention we selected the $k$ most frequent words from the target language, and then we transformed the texts by masking the occurrences of the rest of the words. In another setting, we masked \emph{style-related information} to allow the system to focus only on the topic-related differences between the orientations. We call this a \emph{topic-based model}. For this, we masked the $k$ most frequent words and maintained intact the rest.

After the text transformation by the masking process in both the training and test sets, we represented the documents with character $n$-grams and compared the results obtained with the \emph{style-based} and the \emph{topic-related models}.

\subsection{Experimental Setup}

\begin{description}
\item[Text Transformation:] We evaluated different values of $k$ for extracting the k most frequent words from English\footnote{We use the BNC corpus (https://www.kilgarriff.co.uk/bnc-readme.html) for the extraction of the most frequent words as in \cite{stamatatos2017authorship}.}. For the comparison of the results obtained by each model with the ones of the state-of-the-art, we only showed the results fixing $k=500$.
\item [Text Representation:] We used a standard bag-of-words representation with \textit{tf} weighting and extract character $n$-grams with a frequency lower than 50.
\item [Classifier:] We compared the results obtained with Na\"ive Bayes (NB), Support Vector Machine (SVM) and  Random Forest (RF); for the three classifiers we used the versions implemented in \textit{sklearn} with the parameters set by default.
\item[Evaluation:] We performed 3-fold cross-validation with the same configuration used in \cite{P18-1022}. Therefore, each fold comprised one publisher from each orientation (the classifiers did not learn a publisher's style). We used macro $F_1$ as the evaluation measure since the test set is unbalanced with respect to the three classes. In order to compare our results with those reported in \cite{P18-1022}, we also used accuracy, precision, and recall.
\item[Baseline:] Our baseline method is based on the same text representation with the character n-grams features, but without masking any word.
\end{description}

\subsection{Results and Discussion}

Table \ref{tab:results} shows the results of the proposed method. We compare with \cite{P18-1022} against their topic and style-based methods. In order to compare our
results with those reported in \cite{P18-1022}, we report the same measures the authors used. We also include the macro $F_1$ score because of the unbalance test set. For these experiments we extract the character $5$-grams from the transformed texts, taking into account that as more narrow is the domain more sense has the use of longer n-grams. We follow the steps of \cite{E17-1107} and set k = 500 for this comparison results. The last two rows show the results obtained by applying the system from \cite{P18-1022}\footnote{\url{https://github.com/webis-de/ACL-18}} to our cleaned dataset (Section \ref{mask_hyp}).

\begin{table}[t]
\caption{Results of the proposed masking technique ($k=500$ and $n=5$) applied to mask topic-related information or style-related information. NB: Naive Bayes; RF: Random Forest; SVM: Support Vector Machine. The last two rows show the results obtained by applying the system from \cite{P18-1022} to our cleaned dataset (Section \ref{mask_hyp}).}
\label{tab:results}
\begin{tabular}{|llclccccccccc}
\hline
\multicolumn{1}{|c|}{\multirow{1}{*}{\textbf{Masking}}} & \multicolumn{1}{l|}{\multirow{2}{*}{\textbf{Classifier}}} & \multicolumn{1}{l|}{\multirow{2}{*}{\textbf{Macro F$_1$}}} & \multicolumn{1}{l|}{\multirow{2}{*}{\textbf{Accuracy}}} & \multicolumn{3}{c|}{\textbf{Precision}} & \multicolumn{3}{c|}{\textbf{Recall}} & \multicolumn{3}{c|}{\textbf{F$_1$}} \\
\multicolumn{1}{|c|}{\multirow{1}{*}{\textbf{Method}}} & \multicolumn{1}{l|}{} & \multicolumn{1}{l|}{} & \multicolumn{1}{l|}{} & left & right & \multicolumn{1}{c|}{main} & left & right & \multicolumn{1}{c|}{main} & left & right & \multicolumn{1}{c|}{main} \\ \hline

\multicolumn{1}{|c|}{\multirow{1}{*}{Baseline}} & \multicolumn{1}{l|}{NB} & \multicolumn{1}{c|}{0.52} & \multicolumn{1}{c|}{0.56} & \multicolumn{1}{c|}{0.28} & \multicolumn{1}{c|}{0.57} & \multicolumn{1}{c|}{0.81} & \multicolumn{1}{c|}{\textbf{0.49}} & \multicolumn{1}{c|}{0.58} & \multicolumn{1}{c|}{0.56} & \multicolumn{1}{c|}{0.35} & \multicolumn{1}{c|}{0.57} & \multicolumn{1}{c|}{0.66} \\ \cline{2-13} 
\multicolumn{1}{|c|}{model} & \multicolumn{1}{l|}{RF} & \multicolumn{1}{c|}{0.56} & \multicolumn{1}{c|}{0.62} & \multicolumn{1}{c|}{0.28} & \multicolumn{1}{c|}{0.61} & \multicolumn{1}{c|}{0.80} & \multicolumn{1}{c|}{0.36} & \multicolumn{1}{c|}{0.72} & \multicolumn{1}{c|}{0.63} & \multicolumn{1}{c|}{0.32} & \multicolumn{1}{c|}{0.66} & \multicolumn{1}{c|}{0.70} \\ \cline{2-13} 
\multicolumn{1}{|l|}{} & \multicolumn{1}{l|}{SVM} & \multicolumn{1}{c|}{\textbf{0.70}} & \multicolumn{1}{c|}{\textbf{0.77}} & \multicolumn{1}{c|}{\textbf{0.55}} & \multicolumn{1}{c|}{\textbf{0.75}} & \multicolumn{1}{c|}{\textbf{0.84}} & \multicolumn{1}{c|}{0.42} & \multicolumn{1}{c|}{\textbf{0.79}} & \multicolumn{1}{c|}{\textbf{0.87}} & \multicolumn{1}{c|}{\textbf{0.47}} & \multicolumn{1}{c|}{\textbf{0.77}} & \multicolumn{1}{c|}{\textbf{0.85}} \\ \hline

\multicolumn{1}{|c|}{\multirow{1}{*}{Style-based}} & \multicolumn{1}{l|}{NB} & \multicolumn{1}{c|}{0.47} & \multicolumn{1}{c|}{0.52} & \multicolumn{1}{c|}{0.20} & \multicolumn{1}{c|}{0.51} & \multicolumn{1}{c|}{0.73} & \multicolumn{1}{c|}{0.28} & \multicolumn{1}{c|}{0.65} & \multicolumn{1}{c|}{0.49} & \multicolumn{1}{c|}{0.23} & \multicolumn{1}{c|}{0.57} & \multicolumn{1}{c|}{0.59} \\ \cline{2-13} 
\multicolumn{1}{|c|}{model} & \multicolumn{1}{l|}{RF} & \multicolumn{1}{c|}{0.46} & \multicolumn{1}{c|}{0.53} & \multicolumn{1}{c|}{0.24} & \multicolumn{1}{c|}{0.58} & \multicolumn{1}{c|}{0.64} & \multicolumn{1}{c|}{0.36} & \multicolumn{1}{c|}{0.34} & \multicolumn{1}{c|}{0.73} & \multicolumn{1}{c|}{0.29} & \multicolumn{1}{c|}{0.43} & \multicolumn{1}{c|}{0.68} \\ \cline{2-13} 
\multicolumn{1}{|l|}{} & \multicolumn{1}{l|}{SVM} & \multicolumn{1}{c|}{0.57} & \multicolumn{1}{c|}{0.66} & \multicolumn{1}{c|}{0.33} & \multicolumn{1}{c|}{0.66} & \multicolumn{1}{c|}{0.75} & \multicolumn{1}{c|}{0.26} & \multicolumn{1}{c|}{0.61} & \multicolumn{1}{c|}{0.84} & \multicolumn{1}{c|}{0.29} & \multicolumn{1}{c|}{0.62} & \multicolumn{1}{c|}{0.79} \\ \hline
\multicolumn{1}{|c|}{\multirow{1}{*}{Topic-based}} & \multicolumn{1}{l|}{NB} & \multicolumn{1}{c|}{0.54} & \multicolumn{1}{c|}{0.60} & \multicolumn{1}{c|}{0.26} & \multicolumn{1}{c|}{0.63} & \multicolumn{1}{c|}{0.74} & \multicolumn{1}{c|}{0.36} & \multicolumn{1}{c|}{0.62} & \multicolumn{1}{c|}{0.65} & \multicolumn{1}{c|}{0.29} & \multicolumn{1}{c|}{0.62} & \multicolumn{1}{c|}{0.69} \\ \cline{2-13} 
\multicolumn{1}{|c|}{model} & \multicolumn{1}{l|}{RF} & \multicolumn{1}{c|}{0.53} & \multicolumn{1}{c|}{0.55} & \multicolumn{1}{c|}{0.27} & \multicolumn{1}{c|}{0.64} & \multicolumn{1}{c|}{0.71} & \multicolumn{1}{c|}{0.44} & \multicolumn{1}{c|}{0.60} & \multicolumn{1}{c|}{0.58} & \multicolumn{1}{c|}{0.33} & \multicolumn{1}{c|}{0.61} & \multicolumn{1}{c|}{0.64} \\ \cline{2-13} 
\multicolumn{1}{|l|}{} & \multicolumn{1}{l|}{SVM} & \multicolumn{1}{c|}{0.66} & \multicolumn{1}{c|}{0.74} & \multicolumn{1}{c|}{0.48} & \multicolumn{1}{c|}{0.73} & \multicolumn{1}{c|}{0.81} & \multicolumn{1}{c|}{0.38} & \multicolumn{1}{c|}{0.78} & \multicolumn{1}{c|}{0.82} & \multicolumn{1}{c|}{0.42} & \multicolumn{1}{c|}{0.75} & \multicolumn{1}{c|}{0.82} \\ \hline \hline


\multicolumn{13}{|l|}{\textbf{System from \cite{P18-1022} (applied to our cleaned dataset)}} \\ \hline
\multicolumn{1}{|l|}{Style} & \multicolumn{1}{l|}{RF} & \multicolumn{1}{c|}{0.61} & \multicolumn{1}{c|}{0.63} & \multicolumn{1}{c|}{0.29} & \multicolumn{1}{c|}{0.62} & \multicolumn{1}{c|}{0.71} & \multicolumn{1}{c|}{0.16} & \multicolumn{1}{c|}{0.62} & \multicolumn{1}{c|}{0.80} & \multicolumn{1}{c|}{0.20} & \multicolumn{1}{c|}{0.61} & \multicolumn{1}{c|}{0.74} \\ \hline
\multicolumn{1}{|l|}{Topic} & \multicolumn{1}{l|}{RF} & \multicolumn{1}{c|}{0.63} & \multicolumn{1}{c|}{0.65} & \multicolumn{1}{c|}{0.27} & \multicolumn{1}{c|}{0.65} & \multicolumn{1}{c|}{0.72} & \multicolumn{1}{c|}{0.15} & \multicolumn{1}{c|}{0.62} & \multicolumn{1}{c|}{0.84} & \multicolumn{1}{c|}{0.19} & \multicolumn{1}{c|}{0.63} & \multicolumn{1}{c|}{0.77} \\ \hline


\end{tabular}
\end{table}

Similar to \cite{P18-1022}, the topic-based model achieves better results than the style-related model. However, the differences between the results of the two evaluated approaches are much higher ($0.66$ vs. $0.57$ according to Macro $F_1$) than those shown in \cite{P18-1022}. The highest scores were consistently achieved using the SVM classifier and masking the style-related information (i.e., the topic-related model). This could be due to the fact that all the articles are about the same political event in a very limited period of time. In line with what was already pointed out  in \cite{P18-1022}, the left-wing orientation is harder to predict, possibly because this class is represented with fewer examples in the dataset.

Another reason why our masking approach achieves better results could be that we use a higher length of character n-grams. In fact, comparing the results of \cite{P18-1022} against our baseline model, it is possible to note that even without masking any word, the classifier obtains better results. This suggests that the good results are due to the length of the character n-grams rather than the use of the masking technique.

%
\vspace{1em}
\noindent
\textbf{Robustness of the approach to different values of $k$ and $n$.}
%
%
With the goals of: (i) understanding the robustness of the approach to different parameter 
values; and to see if (ii) it is possible to overcome the $F_1=0.70$ from the baseline model, we vary the values of  $k$ and $n$ and evaluate the macro $F_1$ using SVM.

\begin{figure}[t!]
    \centering
    \begin{subfigure}[t]{0.95\textwidth}
        \centering
        \includegraphics[scale=0.60]{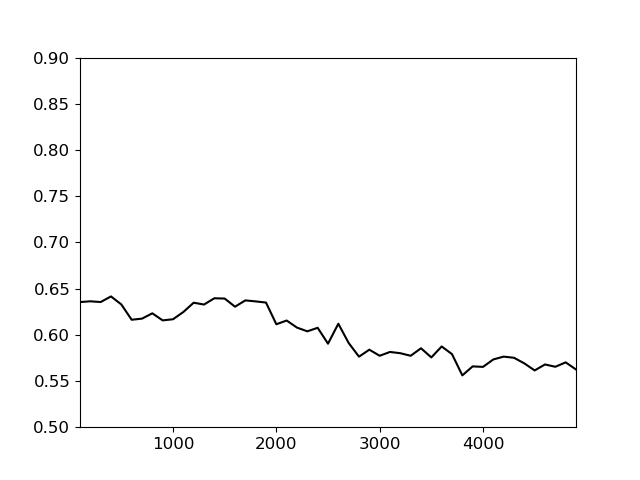}
        \caption{Varying k values and masking the most frequent words: topic-based model.}
    \end{subfigure}%
    ~ 
    
    \begin{subfigure}[t]{0.95\textwidth}
        \centering
        \includegraphics[scale=0.60]{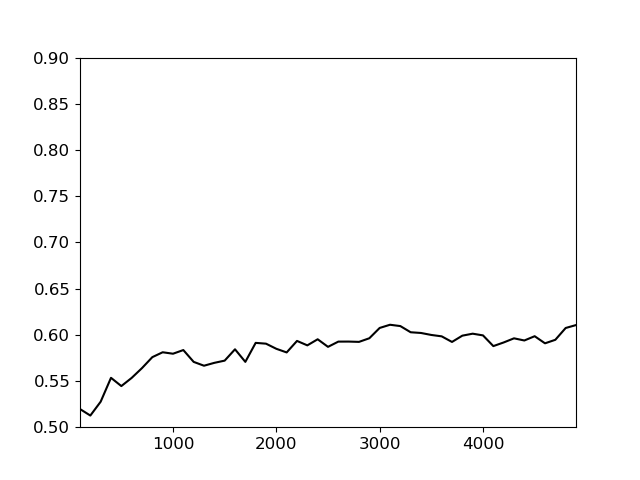}
        \caption{Varying k values and maintaining without masking the most frequent words: style-based-model.}
    \end{subfigure}
  
    \caption{Macro $F_1$ results of the proposed masking technique. We set n=5 for comparing results of different values of $k$.}
    \label{fig:sensitive_k}
\end{figure}

Figures \ref{fig:sensitive_k} shows the results of the variation of $k \in \{100, 200, ..., 5000\}$. When $k>5000$, we clearly can see that the topic-related model, in which the $k$ most frequent terms are masked, is decreasing the performance. This could be explained by the fact that relevant topic-related terms start to be masked too. However, a different behavior is seen in the style-related model, in which we tried to maintain only the style-related words without masking them. In this model, the higher is $k$ the better is the performance. This confirms that for the used dataset, taking into account only style-related information is not good, and observing also topic-related information benefits the classification. When $k$ tends to the vocabulary size, the style-related model tends to behave like the baseline model, which we already saw in Table \ref{tab:results} that achieves the best results.

From this experiment, we conclude that: (i) the topic-related model is less sensitive than the style-related model when $k<500$, i.e. the $k$ most frequent terms are style-related ones; and (ii) when we vary the value of $k$, both models achieve worse results than our baseline.

\begin{figure}[t!]
    \centering
    
    \begin{subfigure}[t]{0.95\textwidth}
        \centering
        \includegraphics[scale=0.6]{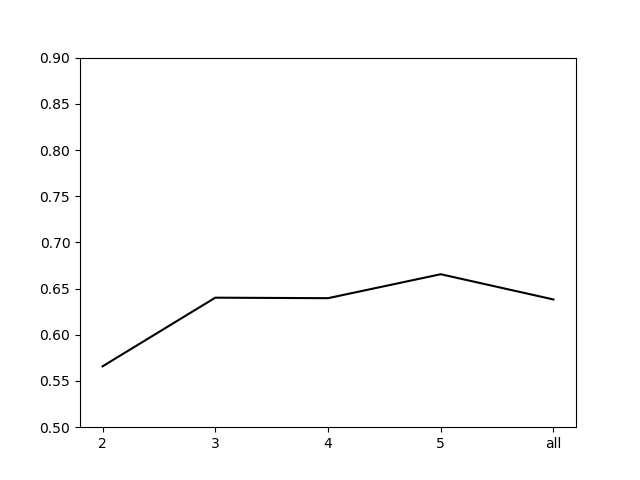}
        \caption{Varying n values and masking the 500 most frequent words.}
    \end{subfigure}
    ~ 
    \begin{subfigure}[t]{0.95\textwidth}
        \centering
        \includegraphics[scale=0.6]{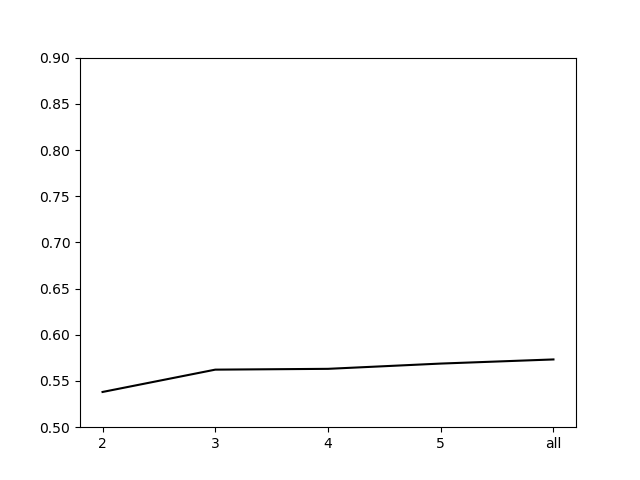}
        \caption{Varying n values and maintaining without masking the 500 most frequent words.}
    \end{subfigure}
    \caption{Macro $F_1$ results of the proposed masking technique. We set n=5 for comparing results of different values of $k$.}
    \label{fig:sensitive_n}
\end{figure}

On the other hand, the results of extracting character $5$-grams are higher than extracting smaller $n$-grams, as can be seen in Figures \ref{fig:sensitive_n}. These results confirm that perhaps the performance of our approach overcomes the models proposed in \cite{P18-1022} because of the length of the $n$-grams\footnote{In \cite{P18-1022} the authors used $n \in [1,3]$.}.

%
\vspace{1em}
\noindent
\textbf{Relevant features.} Table \ref{tab:features} shows the features with the highest weights from the SVM (we use \texttt{scikit-learn}'s method to collect feature weights). It is possible to note that the mention of \textit{cnn} was learned as a discriminative feature when the news from that publisher were used in the training (in the topic-based model). However, this feature is infrequent in the test set where no news from CNN publisher was included.

\begin{table}[]
\centering
\caption{Most relevant features to each class in each model.}
\label{tab:features}
\begin{tabular}{|c|c|c|}
\hline
 \multicolumn{3}{|c|}{\textbf{Baseline model}} \\ \hline 
  \textbf{left} & \textbf{main} & \textbf{right} \\ \hline
 \_imag & \_cnn\_ & e\_are \\ \hline 
  \_that & said\_ & lary\_ \\ \hline
  e\_tru & \_said & \_your \\ \hline
  e\_don & y\_con & n\_pla \\ \hline
  \_here & ry\_co & e\_thi \\ \hline
  
 s\_of\_ & \_cnn\_ & s\_to\_ \\ \hline
  e\_tru & n\_ame & \_your \\ \hline
  for\_h & said\_ & illar \\\hline
  donal & \_said & hilla \\ \hline
  racis & ore\_t & llary \\ \hline
  
 \_here & said\_ & and\_s \\\hline
  \_kill & story & \_hill \\ \hline 
  \_that & \_said & \_let\_ \\ \hline
  \_trum & tory\_ & \_comm \\ \hline
  trump & ed\_be & lary\_ \\ \hline
\end{tabular}~~
\begin{tabular}{|c|c|c|}
\hline
 \multicolumn{3}{|c|}{\textbf{Style-based model}} \\ \hline 
  \textbf{left} & \textbf{main} & \textbf{right} \\ \hline
 but\_* & n\_thi & y\_**\_ \\ \hline 
  out\_w & s\_*\_s & out\_a \\ \hline
  t\_**\_ & \_how\_ & as\_to \\ \hline
  you\_h & at\_he & o\_you \\ \hline
  t\_and & m\_*\_t & ell\_* \\ \hline
 \_is\_a & *\_*\_u & and\_n \\ \hline
  h\_*\_a & e\_\#\_* & hat\_w \\ \hline
  \_of\_\# & and\_* & *\_\#\_\# \\\hline
  or\_hi & **\_*\_ & \_it\_t \\ \hline
  for\_h & t\_the & e\_of\_ \\ \hline
 **\_*\_ & and\_* & o\_you \\\hline
  \_in\_o & \_*\_tw & n\_it\_ \\ \hline 
  hat\_* & *\_two & and\_n \\ \hline
  f\_**\_ & s\_**\_ & \_all\_ \\ \hline
  *\_so\_ & *\_onl & f\_*\_w \\ \hline
\end{tabular}~~
\begin{tabular}{|c|c|c|}
\hline
 \multicolumn{3}{|c|}{\textbf{Topic-based model}} \\ \hline 
  \textbf{left} & \textbf{main} & \textbf{right} \\ \hline
 ant\_* & \_cnn\_ & hilla \\ \hline  
  \_imag & cs\_*\_ & \_*\_da \\ \hline 
  lies\_ & ics\_* & als\_* \\ \hline  
  \_*\_ex & sday\_ & \_*\_le \\ \hline 
  etty\_ & ed\_be & \_dail \\ \hline
 donal & \_cnn\_ & \_*\_te \\ \hline  
  n\_*\_c & day\_* & *\_ame \\ \hline  
  onald & cs\_*\_ & \_*\_am \\ \hline 
  ying\_ & ics\_* & illar \\ \hline 
  thing & *\_*\_e & llary \\ \hline
 â\_*\_* & ed\_be & \_*\_le \\ \hline 
  eâ\_*\_ & y\_con & \_*\_ri \\ \hline  
  nâ\_*\_ & tory\_ & \_hill \\ \hline  
  tâ\_*\_ & story & \_bomb \\ \hline  
  \_imag & d\_bel & *\_*\_r \\ \hline
\end{tabular}
\end{table}

\begin{table}[]
\caption{Fragments of original texts and their transformation by masking the $k$ most frequent terms. Some of the features from Table \ref{tab:features} using the topic-related model are highlighted. }
\label{tab:examples}
\begin{tabular}{|l|p{0.9\textwidth}|}
\hline
\multicolumn{2}{|c|}{\textbf{Topic-related model}} \\ \hline
\multicolumn{1}{|l|}{\multirow{2}{*}{\textbf{left}}} & \multicolumn{1}{p{0.9\textwidth}|}{(...)which his son pr\hl{etty } much confirmed in a foolish statement. The content of those tax returns has been the subject of much speculation, but given Trump’s long history of tax evasion and political bribery, it doesn’t take much \hl{ imag}ination to assume he’s committing some kind of fraud} \\ \cline{2-2} 
\multicolumn{1}{|l|}{} & \multicolumn{1}{p{0.9\textwidth}|}{* * son pr\hl{etty } * confirmed *  foolish statement * content * * tax returns * * * subject * * speculation * * Trump * * * tax evasion * * bribery * doesn * * \hl{ imag}ination * assume * committing * * * fraud} \\ \hline
\multicolumn{1}{|l|}{\multirow{2}{*}{\textbf{main}}} & \multicolumn{1}{p{0.9\textwidth}|}{Obama prov\hl{ed be}yond a shadow of a doubt in 2011 when he released his long-form birth certificate (...) \hl{ CNN } and Fox News cut away at points in the presentation. Networks spent the day talking about Trump's history as a birther (...) Before Friday, the campaign's most recent deception came Wedne\hl{sday } when campaign advisers told reporters that Trump would not be releasing results of his latest medical exam} \\ \cline{2-2} 
\multicolumn{1}{|l|}{} & \multicolumn{1}{p{0.9\textwidth}|}{Obama prov\hl{ed be}yond  shadow * doubt * 2011 * * released * ** birth certificate (...) \hl{ CNN } * Fox News cut * *points * * presentation Networks spent * * talking * Trump * *  birther (...) * Friday * campaign * recent deception * Wedne\hl{sday } * campaign advisers told reporters * Trump * * * releasing results * * latest medical exam} \\ \hline
\multicolumn{1}{|l|}{\multirow{2}{*}{\textbf{right}}} & \multicolumn{1}{p{0.9\textwidth}|}{The email, which \hl{ was da}ted March 17, 2008, and shared with POLITICO, reads: Jim, on Kenya your person in the field might look into the impact there of Obama's public comments about his father. I'm told by State Dept offici\hl{als that} Obama publicly derided his father on (...) Blumenthal, a longtime confidant of both Bill and \hl{Hilla}ry Clinton, emerged as a frequent correspondent in the former secretary of (...)} \\ \cline{2-2} 
\multicolumn{1}{|l|}{} & \multicolumn{1}{p{0.9\textwidth}|}{* email * \hl{ * da}ted March 17 2008 * shared * POLITICO reads Jim * Kenya * * * * field * * * * impact * * Obama * comments * * *  told * * Dept offici\hl{als *}  Obama publicly derided * * (...) Blumenthal  longtime confidant * * Bill * \hl{Hilla}ry Clinton emerged *  frequent correspondent * * former secretary * (...)} \\ \hline

\end{tabular}
\end{table}

Features like \textit{donal} and \textit{onal} are related to Donald Trump, while \textit{illar} and \textit{llary} refer to Hillary Clinton. Each of these names is more frequent in one of the hyperpartisan orientation, and none of them occurs frequently in the mainstream orientation. On the other hand, the relevant features from the style-based model involve function words that are frequent in the three classes (e.g., \textit{out}, \textit{you}, \textit{and}, \textit{of}) even if the combination between function words and other characters can lightly differ in different orientations.

\section{Conclusions}
\label{conclusion}

In this paper we presented initial experiments on the task of hyperpartisan news detection: for this, we explored the use of masking techniques to boost the performance of a lexicalized classifier. Our results corroborate previous research on the importance of content features to detect extreme content: masking, in addition, shows the benefits of reducing data sparsity for this task comparing our results with the state of the art. We evaluated different values of the parameters and see that finally our baseline model, in which we extract character 5-grams without applying any masking process, achieves the better results. As future work we plan to explore more complex learning architectures (e.g., representation learning of masked texts), as well as the application and adaptation of unsupervised methods for detecting ideological positioning from political texts `in the wild' for the online news domain.

\section*{Acknowledgments}
The work of Paolo Rosso was partially funded by the Spanish
MICINN under the research project MISMIS-FAKEnHATE on Misinformation
and Miscommunication in social media: FAKE news and HATE speech
(PGC2018-096212-B-C31).
\bibliographystyle{splncs04}
\bibliography{ms}

\end{document}